\def\year{2022}\relax
\documentclass[letterpaper]{article} % DO NOT CHANGE THIS
\usepackage{aaai22}  % DO NOT CHANGE THIS
\usepackage{times}  % DO NOT CHANGE THIS
\usepackage{helvet}  % DO NOT CHANGE THIS
\usepackage{courier}  % DO NOT CHANGE THIS
\usepackage[hyphens]{url}  % DO NOT CHANGE THIS
\usepackage{graphicx} % DO NOT CHANGE THIS
\usepackage{natbib}  % DO NOT CHANGE THIS AND DO NOT ADD ANY OPTIONS TO IT
\usepackage{caption} % DO NOT CHANGE THIS AND DO NOT ADD ANY OPTIONS TO IT
\DeclareCaptionStyle{ruled}{labelfont=normalfont,labelsep=colon,strut=off} % DO NOT CHANGE THIS
\usepackage{subfigure}
\usepackage{amssymb}
\usepackage{color}
\usepackage{multirow}
\usepackage{booktabs}
\usepackage{algorithm}
\usepackage{algorithmic}
\usepackage{newfloat}
\usepackage{listings}
\floatstyle{ruled}
\newfloat{listing}{tb}{lst}{}
\floatname{listing}{Listing}
\begin{document}

\title{FedProc: Prototypical Contrastive Federated Learning on Non-IID data}

\author {
    % Authors
    Xutong Mu,\textsuperscript{\rm 1}
    Yulong Shen,\textsuperscript{\rm 1}
    Ke Cheng,\textsuperscript{\rm 1}
    Xueli Geng,\textsuperscript{\rm 2}
    Jiaxuan Fu,\textsuperscript{\rm 1}
    Tao Zhang,\textsuperscript{\rm 1}
    Zhiwei Zhang\textsuperscript{\rm 1}
}
\affiliations {
    % Affiliations
    \textsuperscript{\rm 1} School of Computer Science and Technology, Xidian University, Xi'an, China \\
    \textsuperscript{\rm 2} School of Artificial Intelligence, Xidian University, Xi'an, China\\
    % \textsuperscript{\rm 3} State Key Laboratory of Integrated Service Networks (ISN), Xidian University, Xi'an, China\\

    xtmu@stu.xidian.edu.cn, ylshen@mail.xidian.edu.cn, kechengstu@gmail.com, xlgeng@stu.xidian.edu.cn, jxfu\_2099@stu.xidian.edu.cn, taozhang@xidian.edu.cn, zwzhang@xidian.edu.cn 
}

\maketitle

\begin{abstract}
Federated learning allows multiple clients to collaborate to train high-performance deep learning models while keeping the training data locally. However, when the local data of all clients are not independent and identically distributed (i.e., non-IID), it is challenging to implement this form of efficient collaborative learning. Although significant efforts have been dedicated to addressing this challenge, the effect on the image classification task is still not satisfactory. In this paper, we propose \textit{FedProc}: prototypical contrastive federated learning, which is a simple and effective federated learning framework.
% feature learning is innovatively introduced into federated learning, and
The key idea is to utilize the prototypes as global knowledge to correct the local training of each client.
We design a local network architecture and a global prototypical contrastive loss to regulate the training of local models, which makes local objectives consistent with the global optima. Eventually, the converged global model obtains a good performance on non-IID data.
%and a cross-entropy loss, which ensembles the knowledge provided by each client for more effective learning.
%enables the samples from different clients of the same class to be pulled together and samples from different classes to be separated in the embedding space.
%a global prototypical contrastive loss for federated learning is designed,
% pulling the samples from the same class together in the normalized embedding space and pushing the samples from different classes apart
% In this framework, the server obtains the global class-prototypes by gathering the class-prototypes of the client, which is then broadcasted to clients so that the local objective of each client consistent with the global optima.
%the federated averaging of each participant's class-prototype is used to guide the training of feature learning to correct local training.
% Experiments demonstrate that FedProc is significantly better than other state-of-the-art federated learning algorithms on various image classification tasks, and the accuracy is improved by 1.6\% to 7.9\% under the same conditions.
% Extensive empirical results on CIFAR-10, CIFAR-100, and ImageNet demonstrate that FedProc significantly outperforms the state of the art in terms of both inference accuracy and  computational efficiency. For instance, the accuracy is improved by 1.6\% to 7.9\%.
Experimental results show that, compared to state-of-the-art federated learning methods, FedProc improves the accuracy by $1.6\%\sim7.9\%$ with acceptable computation cost.

\end{abstract}

\section{Introduction}
\label{sec:introduction}
Federated learning (FL), as a promising machine learning approach, has enabled distributed clients to collaboratively train a global model without accessing their data by sharing their local model parameters for aggregation.
This approach effectively mitigates privacy concerns in situations where raw data cannot be gathered into a central server for legal or privacy reasons. Serving as a communication-efficient and privacy-preserving learning scheme, FL has shown its potential to facilitate real-world applications, including medical image analysis \cite{kaissis2020secure,kumar2021blockchain}, biometrics analysis \cite{aggarwal2021fedface}, and object detection \cite{liu2020fedvision}, etc.

FL has been shown to work well on the independent and identically distributed (IID) data. %that is, the private data of all clients are distributed in a similar fashion. 
However, in practice, the data hold by different clients usually has a highly skewed distribution. % and even belongs to different data classes. 
Specifically, the local dataset of each client is non-independent and identically distributed (non-IID), which can result in a significant decrease in the performance of FL \cite{zhao2018federated,kairouz2019advances}. This unbalanced data distribution will bring about a drift of local model training in each client, making the local objective far from the global optima. How to mitigate the adverse effects of non-IID data for FL is still an open question.

% A variety of efforts have been made to stabilize local training, by regulating the deviation of local models from a global model over the parameter space.
% FedProx \cite{li2018federated} directly restricts local updates by $l_2$-norm distance, while SCAFFOLD \cite{karimireddy2019scaffold} corrects local updates through variance reduction \cite{johnson2013accelerating}.
% MOON \cite{li2021model} uses contrastive learning in model-level to correct the local updates. This correction is done by minimizing the diversity between the representation learned by the current local model and the representation learned by the global model.
% However, these approachs do not make full use of global information to guide local model training.
% As we show in the experiments (see Section $4$), these approaches fail to achieve good performance on the image classification task.

A variety of efforts have been made to tackle non-IID data issues, mainly from two complementary perspectives: one aims to improve the efficacy of model aggregation, such as FedNova \cite{NEURIPS2020_564127c0}, FedMA \cite{Wang2020Federated}, FedAvgM \cite{hsu2019measuring}.
% These methods are orthogonal to our research and potentially can be combined with these techniques as we work on the local training phase.
Another focuses on stabilizing the local training phase by regulating the deviation of the local models from a global model over the parameter space, such as MOON \cite{li2021model}, FedProx \cite{li2018federated}, SCAFFOLD \cite{karimireddy2019scaffold}. However, whether in the model aggregation phase or local training phase, these approaches do not take full advantage of the underlying knowledge provided by each client.
% differentiate
%may not fully leverage the underlying knowledge across local models, whose diversity suggests informative structural differences of their local data.
As shown in the experiments (see Section 4),
%these approaches fail to achieve good performance on the image classification task.
the accuracy and computation efficiency of these approaches still have plenty of room for improvement.

Observing the challenge in the presence of non-IID data and the limitations of the prior arts, in this paper, we propose a prototypical contrastive federated learning framework, dubbed as FedProc. Inspired by the prototypical contrastive learning \cite{li2020prototypical}, we innovatively introduce prototypes into federated learning for fully utilizing the knowledge of each client to correct the local training. A prototype is defined as the mean vectors for the representations in each class \cite{snell2017prototypical}.
%in order to make full use of the knowledge provided by each client, we innovatively introduce the concept of prototype to federated learning.
% Unlike the traditional prototypical contrastive learning \cite{li2021prototypical}, which is an unsupervised representation learning method that bridges contrastive learning with clustering, FedProc conducts prototypical supervised contrastive learning on each client to make the local objective consistent with the global optima.
%Specifically, we design a hybrid local network architecture containing a global prototypical contrastive loss for feature learning and a cross-entropy loss for classifier learning, to make local objectives consistent with the global optima.
%make local objectives consistent with the global optima.
Specifically, the server first obtains the global class-prototypes by gathering the class-prototypes of the client and broadcasts them to clients as global knowledge to correct local training. Then, the clients use our elaborate local network architecture and loss function to regulate the training of local models, which makes local objectives consistent with the global optima.
This approach forces each sample of the client to be pulled toward the global prototype of its class and pushed away from the global prototypes of other classes, such that the classification performance of the local network would be improved.
%force differently augmented views of each sample to be close to the global prototype of their class and far away from the prototypes of the remaining classes.
In summary, FedProc is a simple and effective federated learning framework that addresses the non-IID data issues from a new perspective of prototype-based contrastive learning.

We experimentally evaluate the performance of FedProc on multiple image classification datasets, including CIFAR-10, CIFAR-100, and Tiny-ImageNet.
FedProc significantly outperforms the state-of-the-art federated learning algorithms \cite{li2021model, li2018federated, karimireddy2019scaffold}.
With acceptable computation cost, FedProc improves accuracy by 1.6\% on the CIFAR-10 dataset, and even more than 7\% on the CIFAR-100 and Tiny-ImageNet datasets. As a highlight, on the CIFAR-100 dataset with 100 clients, FedProc achieves 70.6\% top-1 accuracy, while the best result of existing approaches is 61.8\%.

We summarize our contributions as follows:
\begin{itemize}
\item We propose a novel federated learning framework to address the non-IID data issues. The framework introduces the global class-prototypes to correct the local training, yielding a good classification performance.
%make local objective consistent with the global optima, so that the  performance. boost correct local training so that
\item We design a hybrid local network architecture and a global prototypical contrastive loss to make use of the underlying knowledge provided by global class-prototypes. The careful designs of the local network and the loss function enable FedProc to achieve a good performance.
%being composed of a global prototypical contrastive loss and a cross-entropy loss. The elaborate loss function is more effective for local training.

%ensembling the knowledge of multiple local models over the latent space, which is more effective for learning and communication

%learn better features for improving local training performance. Compared with the conventional contrastive loss, it reduces the computational cost while improving the inference accuracy.
\item We implement FedProc, and do extensive experiments on different datasets. The results demonstrate that FedProc significantly outperforms the state-of-the-art in terms of both inference accuracy and computational efficiency.
\end{itemize}

% Our code is publicly available at github.

\section{Background and Related Work}
\label{section:Relatedwork}

\subsection{Federated Learning}

\begin{figure}[t]
%\vspace{-0.2cm}
\centering
\includegraphics[width=3.1in]{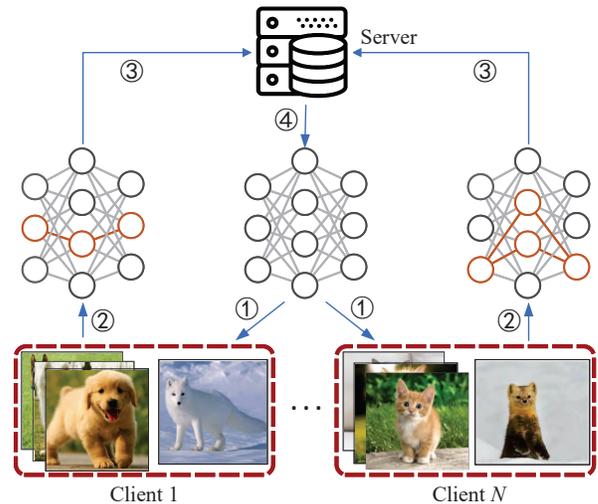}
\caption{Federated learning process.}
\label{fig_model}
\vspace{-0.5cm}
\end{figure}

Federated Learning (FL) is first proposed as a decentralized machine learning paradigm \cite{mcmahan2017communication}, which is executed by following a typical four-step protocol illustrated in Figure \ref{fig_model}. 1) The server randomly initializes the parameters of the global model and sends them to each client. 2) When receiving the global model, each client updates the model based on their local training data using stochastic gradient descent (SGD). 3) The selected clients upload their local model parameters back to the server. 4) The server averages the model parameters to produce a global model for the training of the next round. These steps are repeated until convergence is achieved.

Subsequent work along this line tackles different challenges faced by FL, including heterogeneity \cite{sattler2019robust,briggs2020federated,huang2021personalized}, privacy \cite{truex2019hybrid,wang2019beyond}, communication efficiency \cite{luping2019cmfl,asad2021evaluating,bouacida2021adaptive}, and convergence analysis \cite{huang2021fl,jin2020resource}.
Specifically, a wealth of work has been proposed to handle the non-IID issues, mainly from two complementary perspectives: one focuses on stabilizing the local training phase, such as MOON \cite{li2021model}, FedProx \cite{li2018federated}, SCAFFOLD \cite{karimireddy2019scaffold}. Another aims to improve the efficacy of model aggregation, such as FedNova \cite{NEURIPS2020_564127c0}, FedMA \cite{Wang2020Federated}, FedAvgM \cite{hsu2019measuring}. In addition, there are also other FL studies related to non-IID data setting, such as personalizing the local models for each client \cite{t2020personalized,fallah2020personalized,hanzely2020lower} and designing robust algorithms against different combinations of local distributions \cite{deng2020distributionally,mohri2019agnostic,reisizadeh2020robust}.

\subsection{Contrastive Learning}
Contrastive learning has shown great promise in unsupervised representation learning \cite{chen2020simple,he2020momentum}.
The key idea is to learn an embedding space where samples from the same instance are pulled closer (i.e., positive pairs) and samples from different instances are pushed apart (i.e., negative pairs).
Supervised contrastive learning \cite{khosla2020supervised} is an extension to contrastive learning by incorporating the label information to compose positive and negative images. A recent work \cite{wang2021contrastive} improved the quality of learning features using supervised contrastive learning to solve the long-tail distribution problem in classification tasks.
Later, there emerges prototypical contrastive learning \cite{li2020prototypical}, which is an unsupervised feature learning method that bridges contrastive learning with clustering.
Different from prior work,
%FedProc addresses the non-IID data issue from a perspective of supervised contrastive learning. Moreover,
we design a local network architecture and a loss function tailored for federated learning to address the non-IID data issues from a perspective of supervised contrastive learning.

%by ensembling the knowledge of multiple local models.

% over the feature space, which is more effective for learning and communication
\begin{figure*}[t]
% \vspace{-0.4cm}
\centering
    \subfigure[Client $C_1$ in SOLO]{
        \label{Client1}
        \includegraphics[width=1.45in]{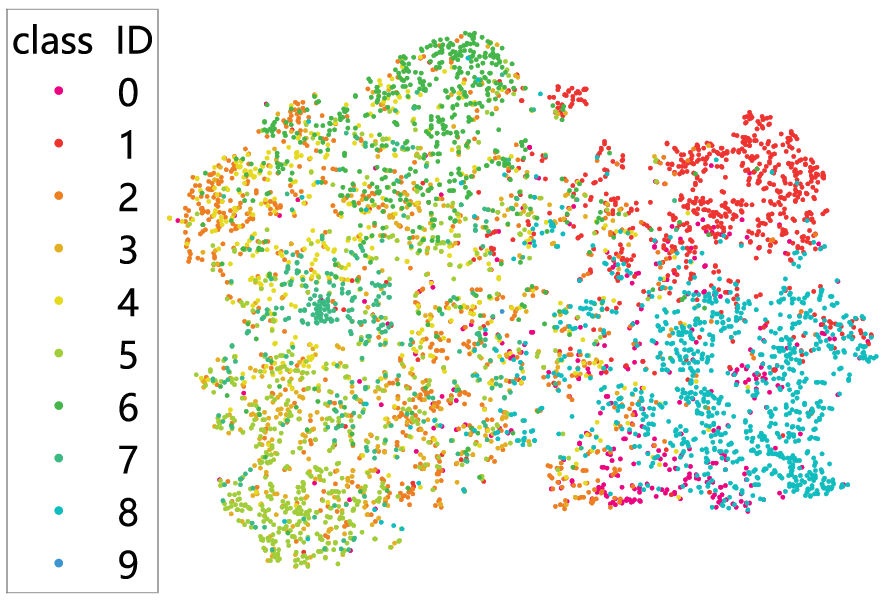}}
    \vspace{0cm}
    \subfigure[Client $C_2$ in SOLO]{
        \label{Client2}
        \includegraphics[width=1.3in]{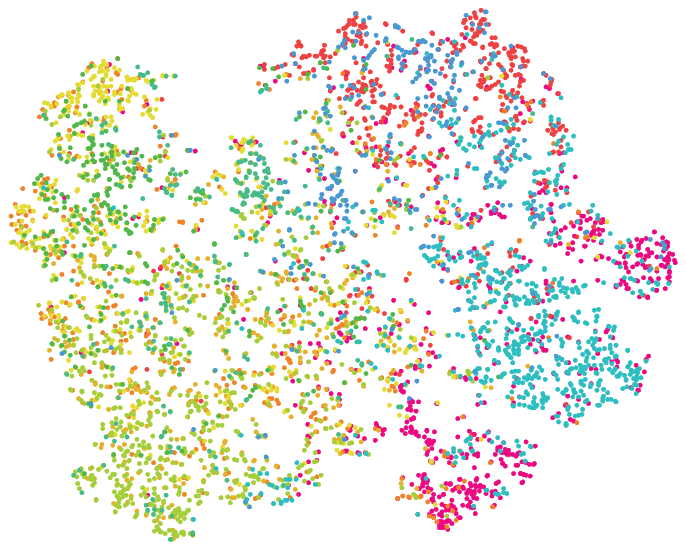}}
        \vspace{0cm}
    \subfigure[Global distribution]{
        \label{global}
        \includegraphics[width=1.3in]{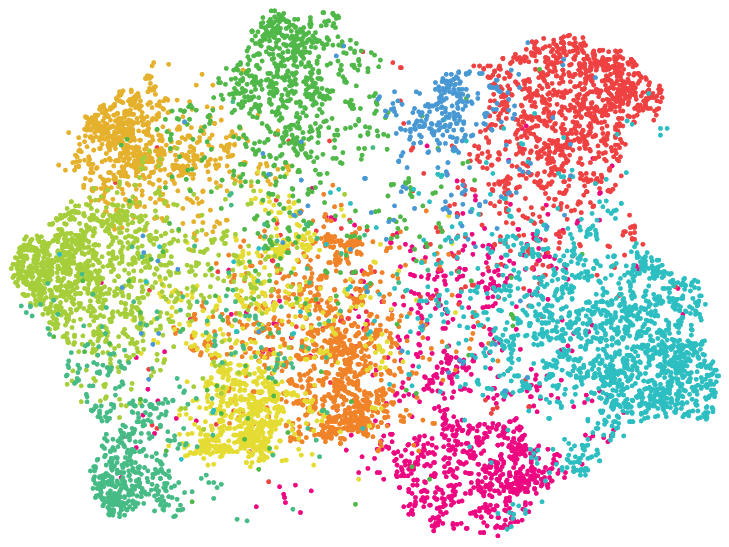}}
        \vspace{0cm}
    \subfigure[Client $C_1$ in FedProc]{
        \label{Client1proc}
        \includegraphics[width=1.3in]{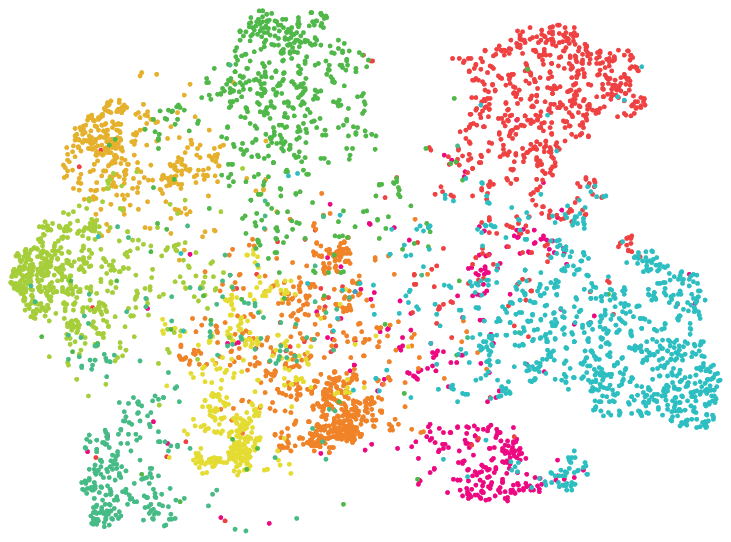}}
        \vspace{0cm}
    \subfigure[Client $C_2$ in FedProc]{
        \label{Client2proc}
        \includegraphics[width=1.3in]{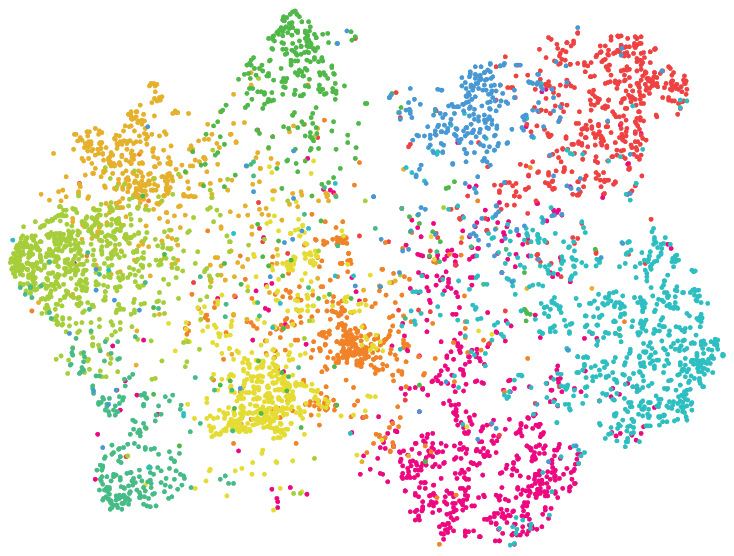}}
\caption{T-SNE visualizations of hidden vectors on CIFAR-10. Figure (a) and (b) show the SOLO representations at Client $C_1$ and $C_2$, respectively. Figure (c) shows global representation distribution. Figure (d) and (e) show the FedProc representations at Client $C_1$ and $C_2$, respectively. SOLO: A baseline approach where each client trains a model only by inputting its local data without federated learning.}
\label{fig_party}
\vspace{-0.4cm}
\end{figure*}
\subsection{Contrastive Learning in Federated Learning}
Contrastive learning in federated learning has recently emerged as an effective approach to tackle the non-IID issue.
Some existing works \cite{zhang2020federated,van2020towards} focus on the unsupervised learning setting.
They use a contrastive loss to compare the representations of different images in order to make full use of the enormous unlabeled data on distributed edge devices.
MOON \cite{li2021model} is based on the design of model-based comparative learning to solve non-IID data problems. This model-level comparative learning is performed by comparing the representations of different model learning, and the local update is corrected by maximizing the consistency between the current local model learning representation and the global model learning representation.
However, in this paper, we focus on the supervised learning setting, and we introduce prototypes to address the issues of inconsistency in the embedding space for each client.

\section{Prototypical Contrastive Federated Learning}

% In this section, we introduce a prototypical contrastive federated learning, which aims to collaboratively train models for a set of clients using the non-IID private data of all clients. in a privacy-preserving way.

% The canonical federated learning problem involves learning a single, global statistical model from data stored on tens to potentially millions of remote devices. In particular, the goal is typically to minimize the following objective function:
% \begin{equation}
% 	\min_w F(w), where\ F(w) = \sum_{k=1}^m p_kF_k(w)
% \end{equation}
% Here $m$ is the total number of devices, $F_k$ is the local objective function for the $k$th device, and $p_k$ specifies the relative impact of each device with $p_k \geq 0$ and $\sum_{k=1}^m p_k=1$.

\subsection{Problem Statement}
FedProc involves $m$ clients (denoted as $C_1,...,C_m$), where $C_i$ has a local dataset $\mathcal{D}_i = {\{(x_j, y_j)\}}^{N^{(i)}}_{j=1}$, where $x_j \in \mathbb{R}^P$ is the $P$-dimensional feature vector of a sample, $y_j \in {1,2, . . . , K}$ (a multi-classification learning task) is the corresponding label of $x_j$, and $N^{(i)}$ is the sample number in dataset $\mathcal{D}_i$.
%={(x_j, y_j)} where $x_j \in \mathbb{R}^D$ is an instance space and $y_j \in \mathbb{R}$ be an output space.
Our goal is to learn a machine learning model $w$ over the dataset
$\mathcal{D} \triangleq \bigcup_{i\in |N|} \mathcal{D}_i$ with the help of a central server, while the raw data are not exchanged. The objective is to solve
% In particular, the goal is typically to minimize the following objective function:

\begin{equation}
	\mathop{\arg\min}\limits_{w} \mathcal{L}(w) = \sum_{i=1}^N \frac{|\mathcal{D}_i|}{|\mathcal{D}|} L_i(w)
\end{equation}
where $L_i(w) = \mathbb{E}_{(x,y)\sim \mathcal{D}_i}[\ell_i(w;(x,y))] $ is the empirical loss of $C_i$, and $\ell_i(w;(x,y))$ is the
loss function.

\subsection{Motivation}
We now discuss the observations that motivate the correction of local training.
We begin by investigating the feature distribution of hidden layers of local network architecture during the training.
For that, we give a baseline approach named SOLO, where each client trains a model only by inputting its local data without federated learning. Specifically, we use SOLO to train models based on the different clients' local data that are both the skewed subsets of CIRFAR-10. Then, we use t-SNE \cite{van2008visualizing} to visualize the hidden layers' features of these local data from two different clients $C_1$ and $C_2$, as shown in Figure \ref{Client1} and Figure \ref{Client2}.
We observe that the feature distributions of images from two clients are quite different in terms of the cluster center and clustering degrees, as well as highly different from the global distribution shown in Figure \ref{global}.
%and the image representations of the same category are scattered in the space.
%For example, the representation vectors of the Class $1$ in Client 1 are distributed in the top-right corner and, while
%is not in the same region as that of the dog in other clients, and may even be in the same region as the cat in other clients.
%Due to the complexity and diversity of local datasets,
As a result, the local objective of each client is inconsistent with the global optima, which can influence the accuracy of federated learning a lot.

%conducted a simple experiment on CIFAR-10. We selected two clients with different data distribution and used t-SNE \cite{van2008visualizing} to visualize the feature of the images in the test data set.
% Federated learning has very low performance on non-IID data because of the drift of the models trained by all clients.
%When the distribution of each local dataset is highly different from the global distribution, the local objective of each client is inconsistent with the global optima.
%According to the feature of hidden layer output in the training process, these clients are not trained in the unified feature space, which may conflict with other clients in the feature space. For example, the representation of the dog in one client is not in the same region as that of the dog in other clients, and may even be in the same region as the cat in other clients.

FedProc tackles the above problem based on an intuitive idea: the prototypes can serve as global knowledge to correct the local training in federated learning. This idea enables the clients to pull the samples from the same class toward the global prototype of its class and away from the global prototypes of other classes, thus making the local objective of each client consistent with the global optima. To demonstrate the efficacy of this idea, we run the FedProc on the above local data of clients $C_1$ and $C_2$, and show the feature distributions of images in Figure \ref{Client1proc} and Figure \ref{Client2proc}.
%we train CNN models using FedProc on the above local data of $C_1$ and $C_2$. As shown in Figure XX,
We find that the points with the same class in the two clients are constrained to the same domain centered in the global class-prototype.
Moreover, the distribution of the points in clients $C_1$ and $C_2$ both match with the global distribution shown in Figure \ref{global}.

\subsection{Method}
% \subsubsection{\textbf{Overview}}
% Based on the above intuition, we propose Fedproc. Fedproc is designed as a simple and effective method based on FedAvg.
% Based on the above intuition, we propose FedProc, a simple and effective method based on FedAvg \cite{mcmahan2017communication}. Due to the drift of local training and the inconsistency of feature space of all clients, FedProc aims to unify the local objectives with the global optima. Inspired by the Prototypical Networks \cite{snell2017prototypical}, we unify the feature space of the client through the global class prototype.
Using the above insight, we present FedProc, a simple and effective FL framework based on FedAvg \cite{mcmahan2017communication}.
Our main changes happen in the local training phase, where the local network architecture and the loss function are carefully designed for learning better representations, which boosts the classification performance of the local network. The overall federated learning algorithm is shown in Algorithm \ref{alg:algorithm}. In the following, we present the local network architecture, the local objectives, and global prototypical contrastive loss.

\begin{figure*}[t]
%\vspace{-0.2cm}
\centering
\includegraphics[width=6.9in]{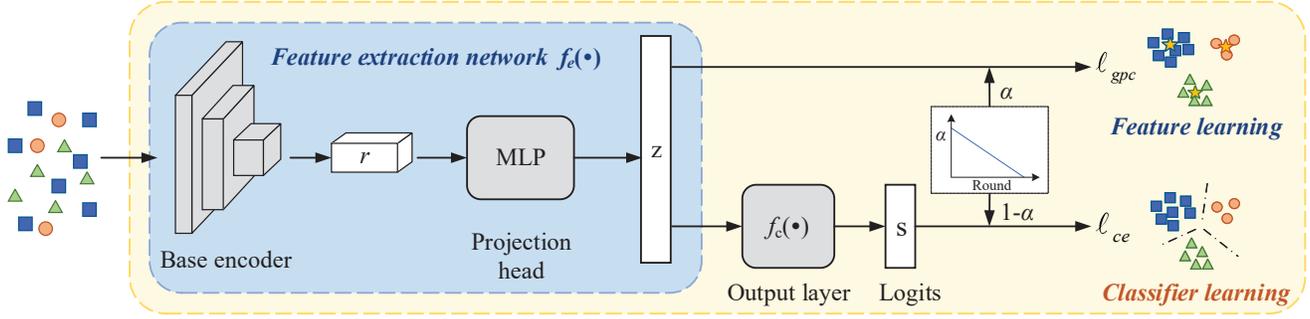}
\caption{Overview of the local network architecture in FedProc. The feature extraction network (including the base encoder and the projection head) extracts the representation $z$, which is used to calculate the global prototypical contrastive loss $\ell_{gpc}$.
By inputting the representation $z$, the output layer $f_c(\cdot)$ predicts the class-wise logits $s$, which are used to compute the cross-entropy loss $\ell_{ce}$. A coefficient $\alpha$ is introduced to adjust the weights of the two loss functions during the local training.}
\label{local_net}
\vspace{-0.5cm}
\end{figure*}

%The local network comprises three modules: a base encoder, a projection head, and an output layer. The representation $z$ extracted by feature extraction network $f_e(\cdot)$ is shared by two loss items. On the one hand, $z$ is used to calculate the global prototypical
%contrastive loss, on the other hand, $z$ is used as the input of the output layer $f_c(\cdot)$ to predict the classification logits. A coefficient $\alpha$ is introduced to adjust the weights of the two terms during the local training phase.

\subsubsection{\textbf{Local Network Architecture}}
Figure \ref{local_net} describes the overview of the proposed local network architecture. The local network is comprised of three modules: a base encoder, a projection head, and an output layer. Firstly, the base encoder extracts representation $r$ from input $x$. Then, the projection head maps the representation $r$ into a vector representation $z \in \mathbb{R}^Q$, which is used to compute a global prototypical contrastive loss $\ell_{gpc}$ (will be illustrated in Eq. (\ref{eq:lgpc})). Noth that, we use a multiple-layer perception (MLP) with one hidden layer to implement the projection head,
%is implemented by a nonlinear multiple-layer perceptron (MLP) with one hidden layer,
which is helpful in improving the representation ability of the layer before it \cite{chen2020simple}. At last, by inputting the image representation $z$,
%In addition, the $l_2$ normalization is applied to $z$ in order that inner product can be used as cosine distance measurements.
%the base encoder first extracts representation $r$ from input $x$, then the projection head maps the representation $r$ into a vector representation $z \in \mathbb{R}^Q$. Last,
the output layer (i.e., a single linear layer $f_c(\cdot)$) predicts the class-wise logits $s \in \mathbb{R}^K$, which are used to compute the cross-entropy loss $\ell_{ce}$.
%produces predicted values for each class.

For ease of presentation, with model weight $w$, we use $w_e$ to represent the weight of the feature extraction network, which is composed of the base encoder and the projection head, and $w_c$ to represent the weight of the output layer.
Correspondingly, $f_e(w_e;\cdot):\mathbb{R}^P \rightarrow \mathbb{R}^Q$ (with learnable parameters $w_e$) represents the feature extraction network, and $f_c(w_c;\cdot):\mathbb{R}^Q \rightarrow \mathbb{R}^K$ (with learnable parameters $w_c$) represents the output layer network.
That is, $z = f_e(w_e;x)$ is the mapped representation of input $x$, and $s = f_c(w_c;z)$ is the prediction vector of the representation $z$.

\renewcommand{\algorithmicrequire}{\textbf{  Input:}}  % Use Input in the format of Algorithm
\renewcommand{\algorithmicensure}{\textbf{  Output:}}  % Use Output in the format of Algorithm

\begin{algorithm}[p]
\caption{The FedProc framework}
\label{alg:algorithm}

\begin{algorithmic}[1] %[1] enables line numbers

	\REQUIRE
	    local datasets $\mathcal{D}_i$, number of communication rounds $T$, number of local epochs $E$, number of classes $K$, number of clients $m$, learning rate $\eta $.
	\ENSURE
	    The final model $w^T$.

\STATE \textbf{\underline{Server executes:}}
	\STATE initialize $w^0$, $c^0$
	\FOR{$t = 0, 1, ..., T-1$}
		\FOR{$i = 1, 2, ..., m$ \textbf{in parallel}}
			\STATE send the global model $w^t$ to $C_i$
			\STATE send the global class-prototypes $c^t$ to $C_i$	
			\STATE $w_i^{t+1}, c_i^{t+1}  \leftarrow$ \textbf{ClientLocalTraining}($i$, $t$, $w^t$, $c^t$)
			% \STATE $c_i^t \leftarrow$ \textbf{LocalClassPrototype}($i$, $w^t$)

		\ENDFOR

		\STATE $c^{t+1} \leftarrow \frac{1}{m}\sum_{i=1}^m c_i^{t+1}$
		% \FOR{$i = 1, 2, ..., m$ \textbf{in parallel}}
		% 	\STATE send the global class-prototypes $c^t$ to $C_i$
		% 	\STATE $w_i^{t+1} \leftarrow$ \textbf{ClientLocalTraining}($i$, $t$, $w^t$, $c^t$)
		% \ENDFOR
		
		\STATE $w^{t+1} \leftarrow \sum_{i=1}^m \frac{\left | \mathcal{D}_i \right |}{\left | \mathcal{D} \right |}w_i^{t+1}$

	\ENDFOR
	\STATE reture $w^T$

% \STATE \textbf{\underline{Client $C_i$ executes:}}
	
	% \STATE \textbf{LocalClassPrototype}($i$, $w^t$)\textbf{:}
	% \STATE $(w_{e_i}^t, w_{c_i}^t) \leftarrow w^t$
	% \STATE $c_{i,k}^t \leftarrow \frac{1}{\left | \mathcal{D}_i^k \right |}\sum_{(x_j,y_j)\in \mathcal{D}_i^k} f_e(w_{e_i}^t;x_j)$	
	
	% \STATE $c_i^t \leftarrow \{c_{i,k=1}^t, c_{i,k=2}^t, ...,c_{i,k=K}^t \}$
	% % \State $c_i^t \leftarrow c^t$
	% \STATE reture $c_i^t$ to server
\STATE \textbf{\underline{ClientLocalTraining}}($i$, $t$, $w^t$, $c^t$)\textbf{:}
	\STATE $(w_{e_i}^t, w_{c_i}^t) \leftarrow w^t$
	\FOR{$epoch = 1,2, ..., E$}
		\FOR{each batch $\textbf{b} = \{x_j, y_j\}$ of $\mathcal{D}_i$}
			\STATE $z_j \leftarrow f_e(w_{e_i}^t; x_j)$
			% \STATE $\ell_{gpc} \leftarrow -log\frac{exp(sim(z_j,c_k))}{\sum_{k'}exp(sim(c_k,c_{k'}))}$
			\STATE $\ell_{gpc} \leftarrow -\log\frac{\exp(sim(z_j,c_{i,k}^t))}{\exp(sim(z_j,c_{i,k}^t))+\sum_{k'}\exp(sim(z_j,c_{i,k'}^t))}$
			\STATE $s_j \leftarrow f_c(w_{c_i}^t; z_j)$
			\STATE $\ell_{ce} \leftarrow CrossEntropyLoss(y_j, s_j)$
            \STATE $\alpha \leftarrow 1-\frac{t}{T}$
			\STATE $\ell \leftarrow  \alpha \cdot \ell_{gpc} + (1-\alpha) \cdot \ell_{ce}$
			\STATE $w_i^{t+1} \leftarrow w_i^t - \eta \nabla \ell$

		\ENDFOR
	\ENDFOR
	\STATE $(w_{e_i}^{t+1}, w_{c_i}^{t+1}) \leftarrow w_i^{t+1}$
	\FOR{$k = 1,2, ..., K$}
		\STATE $c_{i,k}^{t+1} \leftarrow \frac{1}{\left | \mathcal{D}_i^k \right |}\sum_{(x_j,y_j)\in \mathcal{D}_i^k} f_e(w_{e_i}^{t+1};x_j)$	
	\ENDFOR
	\STATE $c_i^{t+1} \leftarrow \{c_{i,1}^{t+1}, c_{i,2}^{t+1}, ...,c_{i,K}^{t+1} \}$
	% \State $c_i^t \leftarrow c^t$
	\STATE reture $w_i^{t+1}, c_i^{t+1}$ to server
    % \STATE reture $w_i^{t+1}$ to server

\end{algorithmic}
\end{algorithm}

% \begin{equation}
% \ell_{ce} \leftarrow CrossEntropyLoss(y_j, s_j)
% \end{equation}

\subsubsection{\textbf{Local Objective}}
The loss function of our local network is composed of two parts. The first part is our proposed global prototypical contrastive loss term $\ell_{gpc}$. This term makes the local network learn an embedding space that has the property of intra-class compactness and inter-class separability.
%This term aims to learn a feature space which has the property of intra-class compactness and inter-class separability.
The second part is a typical cross-entropy loss $\ell_{ce}$ for classifier learning, which can be benefited from the above embedding space.
Inspired by cumulative learning \cite{zhou2020bbn}, we introduce a coefficient $\alpha$ for adjusting the weights
of the two terms during the local training phase. Concretely, the number of total communication rounds is denoted as $T$, and the current round is $t$, $\alpha$ is calculated by $\alpha=1-\frac{t}{T}$.
The final loss function for the network is:
\begin{equation}
	\ell =  \alpha \cdot \ell_{gpc} + (1-\alpha) \cdot \ell_{ce}
\end{equation}
This method makes the local learning to be progressively transited from feature learning to classifier learning with the increased rounds. The local objective is to minimize
\begin{small}
\begin{equation}
	\label{local_obj}
	L_i(w) = \mathbb{E}_{(x,y)\sim \mathcal{D}_i}[\alpha \cdot \ell_{gpc}(w_{e}^t;(x,y))+ (1-\alpha) \cdot \ell_{ce}(w^t;(x,y))]
\end{equation}
\end{small}
In the local training, each client updates the model based on their local training data using stochastic gradient descent (SGD), while the objective is defined in Eq. (\ref{local_obj}).

\subsubsection{\textbf{Global Prototypical Contrastive Loss}}
%The prototype $c_k \in \mathbb{R}^Q$ is the mean representation of the samples belonging to its class:
%\begin{equation}
%	c_k = \frac{1}{\left | \mathcal{D}_i^k \right |}\sum_{(x,y)\in \mathcal{D}_i^k} f_e(w_{e_i};x)
%\end{equation}
%where $\mathcal{D}_i^k$ is the data of class $k$ in the $i$-th client.

To make the global class-prototypes serve as the knowledge to correct each client's local training, we
%To simultaneously ensemble the knowledge of multiple local models and mostly make local objective consistent with global optima, we
propose a global prototypical contrastive loss $\ell_{gpc}$. This loss forces each sample of the client to be close to the global prototype of its class and far away from the global prototypes of other classes.
%differently augmented views of each sample to be close to the global prototype of their class and far away from the prototypes of the remaining classes.
We define the global prototypical contrastive loss as
\begin{equation}
\label{eq:lgpc}
	\ell_{gpc} = -\log\frac{\exp(sim(z_j,c_{i,k}))}{\exp(sim(z_j,c_{i,k}))+\sum_{k'}\exp(sim(z_j,c_{i,k'}))}
\end{equation}
where  $sim(z_j,c_{i,k}) = \frac{z_j^\mathrm{T} c_{i,k}}{\left \|z_j\right \|_2 \cdot \left \|c_{i,k}\right \|_2}$ is the cosine similarity, and $z_j$ is the representation extracted by the feature extraction network when inputting $x_j$.
%$z_j$ is the mapped representation of input $x_j$
Noth that, $c_{i,k}$ (resp. $c_{i,k'}$) denotes the mean representation of the samples belonging to class $k$ (resp. other classes except for class $k$) in the client $C_i$.
%representation for the prototype of class $k$.
The prototype $c_{i,k} \in \mathbb{R}^Q$ is formulated as
%the mean representation of the samples belonging to its class:
\begin{equation}
	c_{i,k} = \frac{1}{\left | \mathcal{D}_i^k \right |}\sum_{(x,y)\in \mathcal{D}_i^k} f_e(w_{e_i};x)
\end{equation}
where $\mathcal{D}_i^k$ is the data of class $k$ in the client $C_i$.

%In the local training, each client uses stochastic gradient descent to update the global model with its local data, while the objective is defined in Eq.(\ref{local_obj}).

\section{Experiment}
We implemented FedProc by PyTorch and ran experiments on the machines running Ubuntu 18.04 and equipped with two NVIDIA GeForce RTX 3090 GPUs and an Intel(R) Core(TM) i9-10900K CPU. To demonstrate the superiority of our work, we compare with the state-of-the-art federated learning algorithms, including 1) MOON \cite{li2021model}, 2) FedAvg \cite{mcmahan2017communication}, 3) FedProx \cite{li2018federated}, 4)SCAFFOLD \cite{karimireddy2019scaffold}, and SOLO. Recall that SOLO is a baseline approach where each client trains a model with its local data without federated learning. In the following experiments, unless explicitly stated, all comparisons with the prior arts use reported results from respective papers.

\subsection{Experimental Setup}
We conduct experiments over three standard datasets: CIFAR-10 (60,000 images with 10 classes), CIFAR-100 (60,000 images with 100 classes), and Tiny-ImageNet (100,000 images with 200 classes). For a fair comparison, we use the same modules in the local network for all approaches. As in the previous work \cite{li2021model}, we use a simple CNN model as the base encoder for CIFAR-$10$ and use ResNet-50 \cite{he2016deep} as the base encoder for CIFAR-100 and Tiny-ImageNet. Note that the simple CNN model has two $5\times5$ convolution layers followed by $2\times2$ max pooling and two fully connected layers with ReLU activation.
For all datasets, the projection head consists of a 2-layer MLP with an output size of 256, and the output layer is just a single linear layer. We use Dirichlet distribution to generate the non-IID data distribution as previous studies \cite{Wang2020Federated,yurochkin2019bayesian}. Specifically, we draw $p^k \sim DirN(\beta)$ from a Dirichlet distribution and allocate a $p^k_i$ proportion of the instances of class $k$ to client $C_i$, where $\beta$ is a concentration parameter controlling the identicalness among clients.
Table \ref{table:defcon} lists the default configuration of our work.

\begin{table}[t]
\small
\caption{The default configuration of our work}
\label{table:defcon}
\centering
\setlength{\tabcolsep}{0.13in}{
\begin{tabular}{c|c}
\toprule
    Parameter                               & Default value                                                               \\
\midrule
    Learning rate ($\eta$)                  & \begin{tabular}[c]{@{}c@{}}$0.01$(CIFAR-$10$)\\ $0.1$ (others)\end{tabular} \\
    Batch size ($B$)                        & $64$                                                                        \\
    Number of clients ($m$)                 & $10$                                                                        \\
    Number of communication rounds ($T$)    & $100$                                                                       \\
    Number of local epochs ($E$)            & $10$                                                                        \\
    Concentration parameter ($\beta$)       & $0.5$ \\
\bottomrule
\end{tabular}}
\end{table}

\begin{figure*}[ht]
% \vspace{-0.4cm}
\centering
    \subfigure[CIFAR-10]{
        \label{cifar10}
        \includegraphics[width=2.25in]{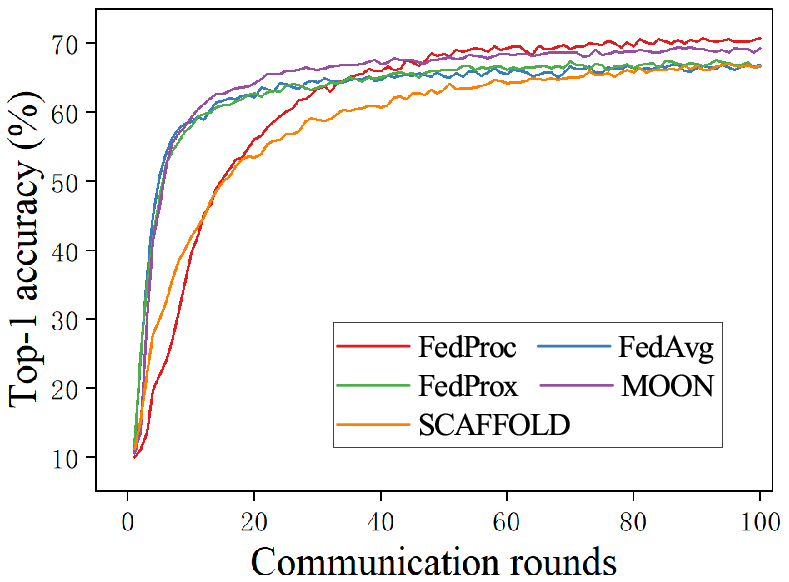}}
    % \hspace{0in}
    \subfigure[CIFAR-100]{
        \label{cifar100}
        \includegraphics[width=2.25in]{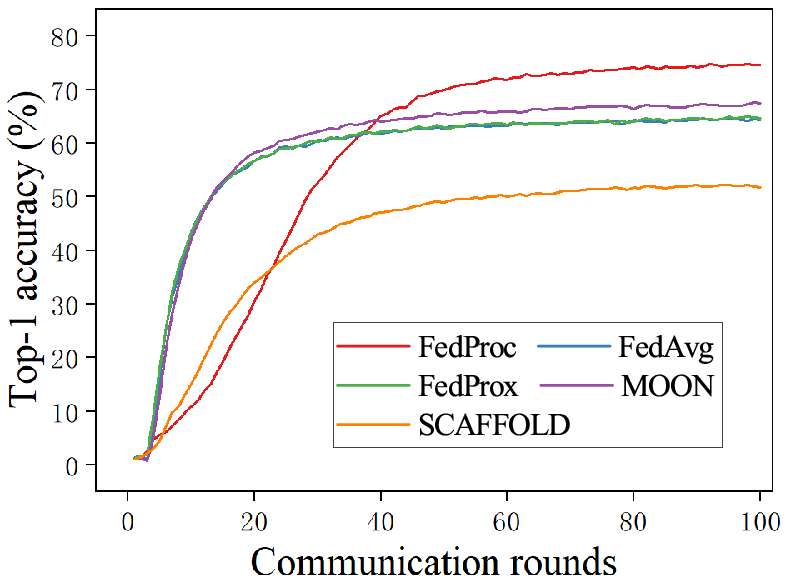}}
    % \vspace{-0.2cm}
    \subfigure[Tiny-ImageNet]{
        \label{ImageNet}
        \includegraphics[width=2.25in]{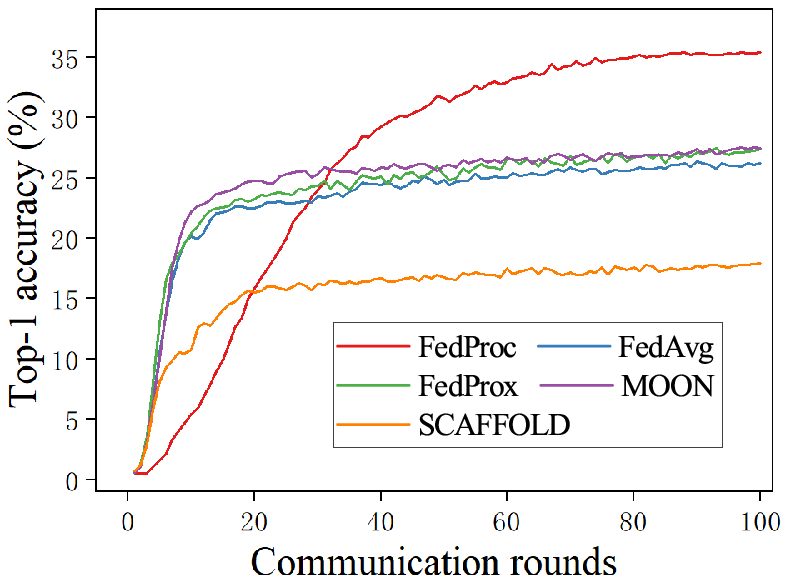}}
\caption{The top-1 test accuracy with different number of communication rounds ($T$).}
\label{Fig5-results}
\vspace{-0.4cm}
\end{figure*}

\begin{figure*}[ht]
% \vspace{-0.4cm}
\centering
    \subfigure[CIFAR-10]{
        \label{cifar10-epoch}
        \includegraphics[width=2.25in]{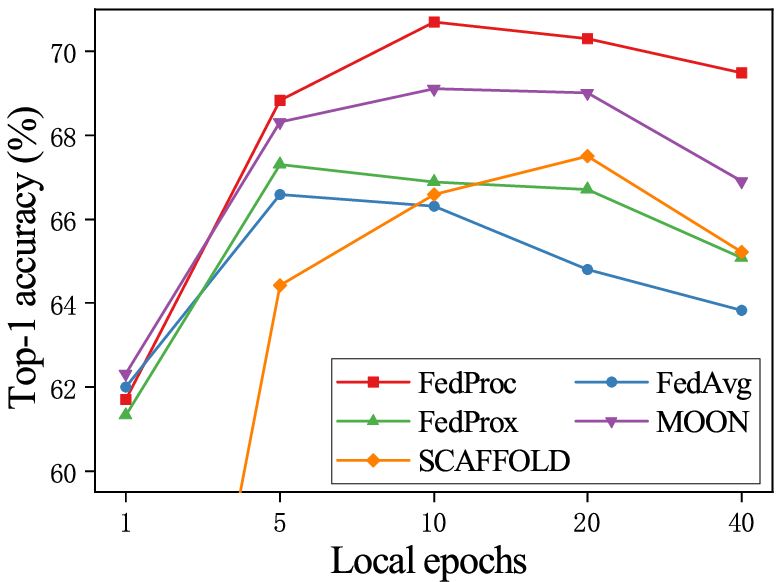}}
    % \hspace{0in}
    \subfigure[CIFAR-100]{
        \label{cifar100-epoch}
        \includegraphics[width=2.25in]{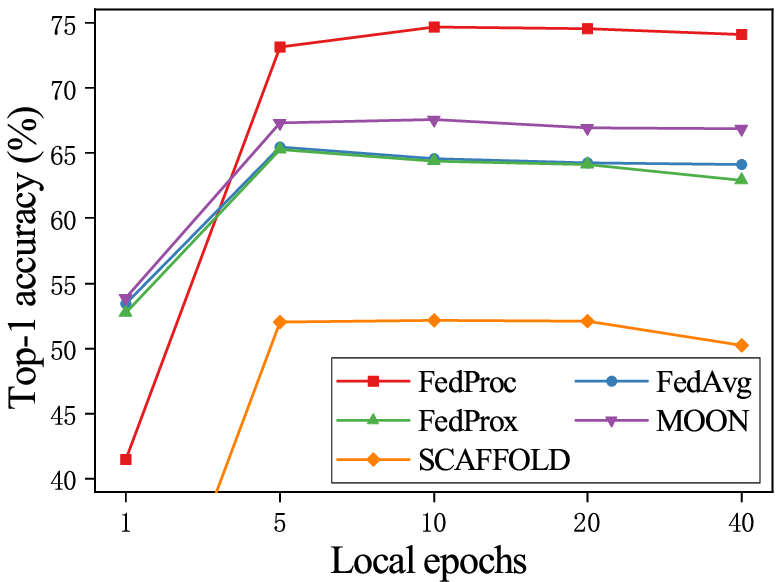}}
    % \vspace{-0.2cm}
    \subfigure[Tiny-ImageNet]{
        \label{ImageNet-epoch}
        \includegraphics[width=2.25in]{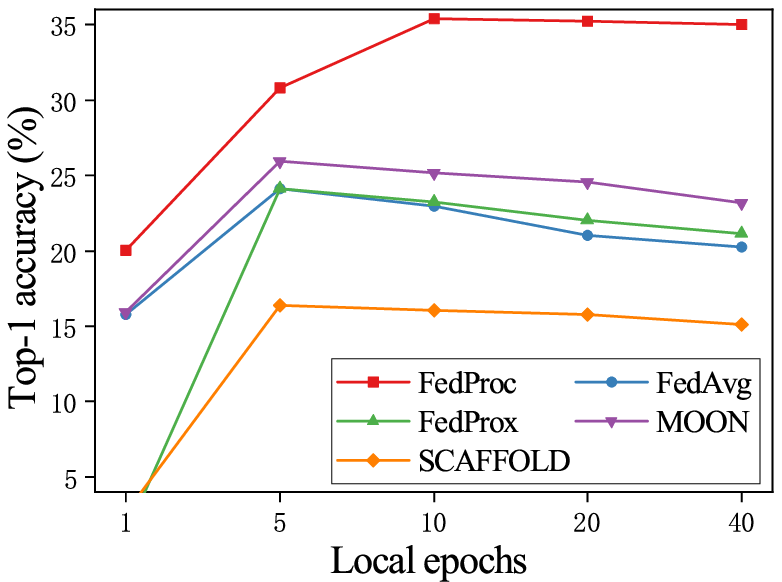}}
\caption{The top-1 test accuracy with different number of local epochs ($E$).}
\label{Fig6-results}
\vspace{-0.4cm}
\end{figure*}

\subsection{Accuracy Results}

Table \ref{table1} lists the top-$1$ test accuracy of all methods. SOLO shows the worst result among all methods, which demonstrates the advantages of federated learning. FedAvg is the first FL framework that uses cross-entropy loss to train the local network, which can be regarded as a baseline for FL. The other FL framework including SCAFFOLD, FedProx, and MOON are designed to address the non-IID data issue. Since FedAvg does not make any optimization for the non-IID setting, the accuracy of FedAvg is relatively low among all FL algorithms. Furthermore, SCAFFOLD is proposed to improve the accuracy on the CIFAR-10, but it has much worse results on CIFAR-100 and Tiny-ImageNet than FedAvg. For FedProx, its accuracy is very close to that of FedAvg. It is because that FedProx makes only minor modifications on the FedAvg by using re-parameterization techniques. MOON presents a model contrastive federated learning, which compares the representations learned by different models. This approach outperforms FedAvg by $1.3\%\sim3\%$ accuracy on the different datasets. As for our method (FedProc), we can observe that its accuracy results are always better than those of other methods for all datasets. Specifically, our method outperforms MOON by $1.6\%\sim7.9\%$ on the different datasets. It is indicated that our method (prototypical contrastive federated learning) can effectively correct the local training. 
%Therefore the accuracy increases remarkably. 
Next, we explore the impact of different parameters on accuracy.

\begin{table}[t]
\small
\caption{The top-$1$ accuracy of FedProc and the other methods on test datasets.}
\label{table1}
\centering
\setlength{\tabcolsep}{0.03in}{
\begin{tabular}{c|c|c|c}
\toprule
    Method   & CIFAR-10                    & CIFAR-100                   & Tiny-ImageNet             \\
\midrule
    SOLO     & $46.3\%\pm5.1\%$            & $22.3\%\pm1.0\%$            & $8.9\%\pm0.3\%$           \\
    FedAvg   & $66.3\%\pm0.5\%$            & $64.5\%\pm0.4\%$            & $26.2\%\pm0.1\%$          \\
    SCAFFOLD & $66.6\%\pm0.2\%$            & $52.5\%\pm0.3\%$            & $17.9\%\pm0.2\%$          \\
    FedProx  & $66.9\%\pm0.2\%$            & $64.6\%\pm0.2\%$            & $27.4\%\pm0.2\%$          \\
    MOON     & $69.1\%\pm0.4\%$            & $67.5\%\pm0.4\%$            & $27.5\%\pm0.1\%$          \\
    FedProc  & $\mathbf{70.7\%\pm0.3\%}$   & $\mathbf{74.6\%\pm0.1\%}$   & $\mathbf{35.4\%\pm0.1\%}$ \\
\bottomrule
\end{tabular}}
\end{table}

\subsubsection{Impact of number of communication rounds ($T$)}
Figure \ref{Fig5-results} shows the accuracy in each round during the training. We find that FedProc achieves the best performance at the end of the training. Further, the curves in Figure \ref{Fig5-results} show that FedProc improves the accuracy at the expense of the slow convergence speed. This is because feature learning plays a critical role at the beginning of training, and then classifier learning gradually dominates the training. In other words, FedProc learns better representations in the early stages of the training, which can benefit the classifier learning in the later stages.

\subsubsection{Impact of number of local epochs ($E$)}
Figure \ref{Fig5-results} shows the accuracy as the number of local epoch increases during the training. We find that the accuracy of  most of the methods is the highest when the number of local epochs $E=10$. This is because that, when $E$ is small, the local network can not be fully trained. But, when $E > 10$, there is over-fitting in the local training on the skewed data, which leads to a reduction in the accuracy of the global model.
%makes the accuracy drop.
%We study the effect of number of local epochs on the accuracy of final model. The results are shown in Figure 5. When the number of local epochs is 1, the local update is very small. Thus, the training is slow and the accuracy is relatively low given the same number of communication rounds. All approaches have a close accuracy. When the number of local epochs becomes too large, the accuracy of all approaches drops, which is due to the drift of local updates(i.e., the local optima are not consistent with the global optima). Nevertheless, FedProc clearly outperforms the other approaches. This further verifies that FedProc can effectively mitigate the negative effects of the drift by too many local updates.

\subsubsection{Impact of data heterogeneity ($\beta$)}
% We study the effect of data heterogeneity by varying the concentration parameter $\beta$ of Dirichlet distribution on CIFAR-100. For a smaller $\beta$, the partition will be more unbalanced. The results are shown in Table \ref{table3}. FedProc always achieves the best accuracy among three unbalanced levels.
% When $\beta = 0.5$ and $\beta = 5$, FedProc is more than 7\% higher than MOON on the CIFAR-100 and Tiny-ImageNet datasets.
% When the unbalanced level decreases (i.e., $\beta = 5$), FedProx is worse than FedAvg, while FedProc still outperforms FedAvg  on CIFAR-10, CIFAR-100 and Tiny-ImageNet datasets with an accuracy of 6\%, 10.1\% and 9.9\%, respectively. The experiments demonstrate the effectiveness and robustness of FedProc.

To assess the impact of the data heterogeneity on the accuracy, we ran the experiments on heterogeneous data by varying the concentration parameter $\beta$ of the Dirichlet distribution on the CIFAR100 dataset. A smaller $\beta$ indicates a more skewed data distribution. The results in Table \ref{table3} shows that FedProc consistently achieves the best accuracy with all levels of imbalance. Specifically, FedProc outperforms MOON by $7.6\%$ accuracy when $\beta = 5$. 
%As the unbalanced level increases ($\beta = 0.5, 0.1$),
When the data distributions are highly heterogeneous ($\beta = 0.5, 0.1$), FedProc still outperformed MOON by $7.1\%$ and $4.9\%$ accuracy, respectively. This result verifies our motivations, since the advantage of FedProc benefits from the introduction of class prototypes, which serve as global knowledge to correct the local training. In contrast, other methods do not make full use of the underlying knowledge, such as the global class-prototypes.
%FEDGEN is induced from the knowledge distilled to local users, which mitigates the discrepancy of latent distributions across users. This knowledge is otherwise not accessible by baselines such as FEDAVG or FEDPROX.

%more than 7\% higher than MOON. When the unbalanced level increases ($\beta = 0.1$), FedProc still outperformed MOON by 4.9\%, with an accuracy of 68.9\%. From the results in Table \ref{table3}, we can find that the improvement of the  non-IID FL algorithms FedProx is insignificant compared with FedAvg, which is smaller than 0.5\%. MOON improved FL in the non-IID case, but also improved accuracy by less than 3\%. FedProc consistently achieves the best accuracy among these three levels of imbalance. When $\beta = 5$, FedProc is more than 7\% higher than MOON. When the unbalanced level increases ($\beta = 0.1$), FedProc still outperformed MOON by 4.9\%, with an accuracy of 68.9\%.

\begin{table}[t]
\small
\caption{The top-1 test accuracy with $\beta=5, 0.5, 0.1$.}
\label{table3}
\centering
\setlength{\tabcolsep}{0.19in}{
\begin{tabular}{c|c|c|c}
\toprule
    Method   & $\beta$ = 5         & $\beta$ = 0.5       & $\beta$ = 0.1      \\
\midrule
    SOLO     & $26.6\%$            & $22.3\%$            & $15.9\%$           \\
    FedAvg   & $65.7\%$            & $64.5\%$            & $62.5\%$           \\
    SCAFFOLD & $55.0\%$            & $52.5\%$            & $47.3\%$           \\
    FedProx  & $64.9\%$            & $64.6\%$            & $62.9\%$           \\
    MOON     & $68.0\%$            & $67.5\%$            & $64.0\%$           \\
    FedProc  & $\mathbf{75.6\%}$   & $\mathbf{74.6\%}$   & $\mathbf{68.9\%}$  \\
\bottomrule

\end{tabular}}
\end{table}

\begin{figure*}[t]
% \vspace{-0.4cm}
\centering
    \subfigure[50 clients]{
        \label{p50}
        \includegraphics[width=3.1in]{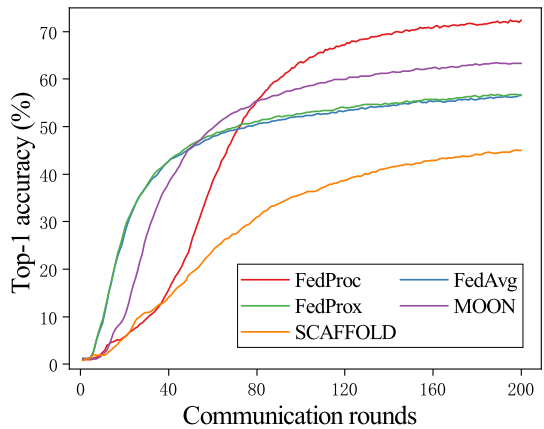}}
        \hspace{5mm}
    \subfigure[100 clients]{
        \label{p100}
        \includegraphics[width=3.1in]{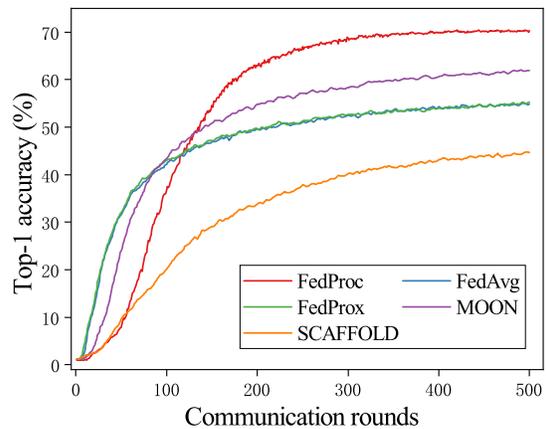}}
\caption{The top-1 test accuracy on CIFAR-100 with 50/100 clients.}
\label{Fig7-results}
\vspace{-0.4cm}
\end{figure*}

\begin{table}[t]
\small
\caption{The top-1 accuracy with different kinds of loss of local objective.}
\label{table5}
\centering
\setlength{\tabcolsep}{0.075in}{
\begin{tabular}{c|c|c|c}
\toprule
    Method                 & CIFAR-10           & CIFAR-100          & Tiny-ImageNet    \\
\midrule
    Two-stage FL           & $65.10\%$          & $67.50\%$          & $24.70\%$          \\
    FedProc($\alpha$= 0.5) & $66.30\%$          & $68.90\%$          & $30.20\%$          \\
    FedProc                & $\mathbf{70.70\%}$ & $\mathbf{74.60\%}$ & $\mathbf{35.40\%}$ \\
\bottomrule
\end{tabular}}
\end{table}

\subsubsection{Impact of coefficient in loss function ($\alpha$)}

% In this work, we use a curriculum to smoothly transit the training from feature learning to classifier learning.
% To justify the advantage of this learning strategy, we firstly designed a two-stage federated learning inspired by two-stage work \cite{khosla2020supervised}, which trains the features using SC loss in the first stage and then fixes the features to train classifiers in the second stage.
% From Table \ref{table5} we can see, this two-stage training scheme results in obviously inferior performance to our curriculum based training, because it harms the compatibility between the features and classifiers. To further highlight the importance of the curriculum, we set the weighting coefficient $\alpha$ to be 0.5. Still, unsatisfactory results are obtained. When the curriculum is used, we allow the supervised contrastive losses to dominate the training first in order to fully exploit their capacity to learn discriminative features, which can benefit the classifier learning in later phase.

In this work, we use the coefficient $\alpha$ to adjust the weights of the feature learning and classifier learning during the local training. To demonstrate the superiority of our method, we design a two-stage federated learning inspired by two-stage work \cite{khosla2020supervised}. This method trains the features by $\ell_{gpc}$ loss in the first stage and then fixes the features to train classifiers in the second stage. As shown in Table \ref{table5}, the accuracy of our method is evidently higher than that of the two-stage training method. It is because that the training with two stages breaks the compatibility between feature learning and classifier learning. To validate the efficacy of the setting method of $\alpha$, we re-run FedProc when fixing $\alpha=0.5$. Obviously, the results in this setting are worse than ours. When $\alpha=1-\frac{t}{T}$, FedProc learns better representations in the early stages of the training, making it have the greater classification capacity in the later stages.
%to control the network training process to smoothly transition from feature learning to classifier learning.
%To justify the advantage of this learning strategy, we firstly designed a two-stage federated learning inspired by two-stage work \cite{khosla2020supervised}, which trains the features using $\ell_{gpc}$ loss in the first stage and then fixes the features to train classifiers in the second stage.
%From Table \ref{table5} we can see, this two-stage training scheme results in obviously inferior performance to our curriculum based training, because it harms the compatibility between the features and classifiers. 
%To further highlight the importance of the curriculum, we set the weighting coefficient $\alpha$ to be 0.5. Still, unsatisfactory results are obtained. When the curriculum is used, 
%we allow the supervised contrastive losses to dominate the training first in order to fully exploit their capacity to learn discriminative features, which can benefit the classifier learning in later phase.

\subsection{Computation Cost}
To make a fair comparison, we measure the computation cost of all the above methods under the same machines. Table \ref{table4} shows the average training time per round. We can observe that the average training time of FedAvg is the lowest of all. The reason is, FedProx, MOON, and FedProc introduce additional loss items based on FedAvg, and SCAFFOLD introduces additional control variables for the server and clients. We also find that the average training time of FedProc on CIFAR-$10$ and CIFAR-$100$ is nearly the same as most of the methods (e.g., SCAFFOLD and FedProx).
As a highlight, FedProc on Tiny-ImageNet is superior to other methods (except FedAvg). We can conclude that FedProc has more advantages in the computation cost as the volume of data and the scale of local networks increases.
% As a highlight, FedProc is superior to other methods (except FedAvg) for Tiny-ImageNet in terms of the average training time. We can conclude that FedProc has more advantages in the computation cost as the volume of data and the scale of local networks increases.
%, so the training of these methods is slower than FedAvg.
%As can be seen from Table 1, the average training time per round of FedProc did not increase drastically along with the increase in the amount of data. Especially on the Tiny-ImageNet dataset, FedProc is superior to other methods except FedAvg.

% Since FedProc introduces an additional loss term in the local training phase, the training of FedProc will be slower than FedAvg. For the experiments in Table \ref{table1}, the average training time per round with two NVIDIA GeForce RTX 3090 GPUs and a Intel(R) Core(TM) i9-10900K CPU are shown in Table \ref{table4}.
% We can see from Table \ref{table4} that the average training time per round of FedProc is slightly longer than that of FedAvg on the three data sets. However, on the datasets CIFAR-100 and Tiny-ImageNet, the average training time per round of FedProc has a significant improvement compared with that of MOON.

% Compared with FedAvg, the computation overhead of FedProc is acceptable
% especially on CIFAR-10 and CIFAR-100.

\begin{table}[t]
\small
\caption{The average training time per round.}
\label{table4}
\centering
\setlength{\tabcolsep}{0.10in}{
\begin{tabular}{c|c|c|c}
\toprule
    Method   & CIFAR-10          & CIFAR-100          & Tiny-ImageNet       \\
\midrule
    FedAvg   & 8min34s           & 24min36s           & 104min11s           \\
    SCAFFOLD & \textbf{8min43s}  & 27min55s           & 181min56s           \\
    FedProx  & 8min52s           & \textbf{27min28s}  & 117min7s            \\
    MOON     & 9min6s            & 45min5s            & 186min10s           \\
    FedProc  & 11min57s          & 27min41s           & \textbf{111min12s}  \\
\bottomrule
\end{tabular}}
\end{table}

\subsection{Scalability}
In order to demonstrate the scalability of FedProc, we ran the experiments on CIFAR-100 with a large number of clients. As in the previous work \cite{li2021model}, the number of clients ($m$) is set to $50$ (with sampling rate $\gamma=1$) and $100$ (with sampling rate $\gamma=0.2$). Note that $\gamma=0.2$ means that $20$ clients out of $100$ clients are randomly selected to participate in the training in each round (refer to FedAvg \cite{mcmahan2017communication} for client sampling technology).
The results in Table \ref{table2} and Figure \ref{Fig7-results} demonstrate the excellent scalability of FedProc, whose accuracy is far higher than those of the other methods. In particular, our method outperforms MOON by $9.3\%$ accuracy when the number of rounds $T=200$ and the number of clients $m=50$.
The excellent scalability of FedProc is due to the introduction of prototypical contrastive learning. This improvement makes local objectives of each client consistent with the global optima, such that the performance of FedProc will not be affected as the number of clients increases.

%we achieve $9.3\%$ and $8.8\%$ accuracy improvements under 200 rounds of 50 clients and 500 rounds of 100 clients, respectively.
% To show the scalability of FedProc, we try a larger number of clients on CIFAR-100. Specifically, we try two settings: (1) We partition the dataset into 50 clients and all clients participate in federated learning in each round. (2) We partition the dataset into 100 clients and randomly sample 20 clients to participate in federated learning in each round (client sampling technique introduced in FedAvg). The results are shown in Table \ref{table2}.
% For FedProc, it outperforms the FedAvg and FedProx over 15.9\% accuracy at 200 rounds with 50 clients and 15.6\% accuracy at 500 rounds with 100 clients. For MOON, we show the results with $\mu = 10$ (best $\mu$ from the previous paper \cite{li2021model}).
% % For FedProc, we show the results with $\mu = 1$ (best $\mu$ from Section 4.2) and $\mu = 10$ .  Moreover, for FedProc ($\mu= 10$), although the large model-contrastive loss slows down the training at the beginning as shown in Figure 8, FedProc can outperform the other approaches a lot with more communication rounds.
% Compared with MOON, FedProc achieves about 8.8\% higher accuracy at 200 rounds with 50 clients and at 500 rounds with 100 clients. SCAFFOLD has a low accuracy with a relatively large number of clients.

\begin{table}[t]
\small
\caption{The top-1 test accuracy with varying number of clients ($m$) and varying number of communication rounds ($T$) on CIFAR-$100$.}
\label{table2}
\centering
\begin{tabular}{c|c|c|c|c}
\toprule
    \multirow{2}{*}{Method} & \multicolumn{2}{c|}{$m=50$}            & \multicolumn{2}{c}{$m=100$}           \\ \cline{2-5}
                            & $T=100$           & $T=200$            & $T=250$       & $T=500$                \\
\midrule
    % SOLO                    & \multicolumn{2}{c|}{$10\%$}            & \multicolumn{2}{c}{$7.3\%$}           \\
    FedAvg                  & $51.9\%$          & $56.4\%$           & $51.0\%$           & $55.0\%$          \\
    SCAFFOLD                & $35.8\%$          & $44.9\%$           & $37.4\%$           & $44.5\%$          \\
    FedProx                 & $52.7\%$          & $56.6\%$           & $51.3\%$           & $54.6\%$          \\
    MOON                    & $58.2\%$          & $63.2\%$           & $56.9\%$           & $61.8\%$          \\
    FedProc                 & $\mathbf{63.6\%}$ & $\mathbf{72.5\%}$  & $\mathbf{68.9\%}$  & $\mathbf{70.6\%}$ \\
\bottomrule
\end{tabular}
\end{table}

\section{Conclusion}
This paper proposes prototypical contrastive federated learning (FedProc), a simple and effective federated learning framework to tackle non-IID data issue. FedProc introduces class prototypes as global knowledge to correct the local training in federated learning. Technically, we design a local network architecture and global prototypical contrastive loss to make local objectives consistent with the global optima, yielding a good classification performance of the global model.
%Under this framework, the prototype is innovatively introduced into federated learning to correct the drift of local training under non-IID settings.
Extensive experiments on multiple datasets demonstrate the advantage of FedProc on non-IID data. 
% have shown that our proposed approach greatly improves the accuracy with acceptable computation, compared with the state-of-the-art.
% In addition, our method is more robust and has higher scalability.

% \section*{Acknowledgment}
% The corresponding authors are Liangmin Wang and Yulong Shen. This research was supported in part by the National Key Research and Development Program of China (Grant No. 2018YFE0207600), the National Natural Science Foundation of China (Grant No. U1736216, 61571352 and 61602364), Key R\&D Program of Shaanxi Province (Grant No. 2019ZDLGY12-03, 2019ZDLGY13-06), the Fundamental Research Funds for the Central Universities and the Innovation Fund of Xidian University.

\bibliography{aaai22}

\begin{thebibliography}{41}
\providecommand{\natexlab}[1]{#1}

\bibitem[{Aggarwal, Zhou, and Jain(2021)}]{aggarwal2021fedface}
Aggarwal, D.; Zhou, J.; and Jain, A.~K. 2021.
\newblock FedFace: Collaborative Learning of Face Recognition Model.
\newblock arXiv:2104.03008.

\bibitem[{Asad et~al.(2021)Asad, Moustafa, Ito, and Aslam}]{asad2021evaluating}
Asad, M.; Moustafa, A.; Ito, T.; and Aslam, M. 2021.
\newblock Evaluating the communication efficiency in federated learning
  algorithms.
\newblock In \emph{2021 IEEE 24th International Conference on Computer
  Supported Cooperative Work in Design (CSCWD)}, 552--557. IEEE.

\bibitem[{Bouacida et~al.(2021)Bouacida, Hou, Zang, and
  Liu}]{bouacida2021adaptive}
Bouacida, N.; Hou, J.; Zang, H.; and Liu, X. 2021.
\newblock Adaptive federated dropout: Improving communication efficiency and
  generalization for federated learning.
\newblock In \emph{IEEE INFOCOM 2021-IEEE Conference on Computer Communications
  Workshops (INFOCOM WKSHPS)}, 1--6. IEEE.

\bibitem[{Briggs, Fan, and Andras(2020)}]{briggs2020federated}
Briggs, C.; Fan, Z.; and Andras, P. 2020.
\newblock Federated learning with hierarchical clustering of local updates to
  improve training on non-IID data.
\newblock In \emph{2020 International Joint Conference on Neural Networks
  (IJCNN)}, 1--9. IEEE.

\bibitem[{Chen et~al.(2020)Chen, Kornblith, Norouzi, and
  Hinton}]{chen2020simple}
Chen, T.; Kornblith, S.; Norouzi, M.; and Hinton, G. 2020.
\newblock A Simple Framework for Contrastive Learning of Visual
  Representations.
\newblock In \emph{International Conference on Machine Learning}, 1597--1607.
  PMLR.

\bibitem[{Deng, Kamani, and Mahdavi(2020)}]{deng2020distributionally}
Deng, Y.; Kamani, M.~M.; and Mahdavi, M. 2020.
\newblock Distributionally Robust Federated Averaging.
\newblock \emph{Advances in Neural Information Processing Systems}, 33.

\bibitem[{Fallah, Mokhtari, and Ozdaglar(2020)}]{fallah2020personalized}
Fallah, A.; Mokhtari, A.; and Ozdaglar, A. 2020.
\newblock Personalized federated learning with theoretical guarantees: A
  model-agnostic meta-learning approach.
\newblock \emph{Advances in Neural Information Processing Systems}, 33:
  3557--3568.

\bibitem[{Hanzely et~al.(2020)Hanzely, Hanzely, Horv{\'a}th, and
  Richtarik}]{hanzely2020lower}
Hanzely, F.; Hanzely, S.; Horv{\'a}th, S.; and Richtarik, P. 2020.
\newblock Lower Bounds and Optimal Algorithms for Personalized Federated
  Learning.
\newblock \emph{Advances in Neural Information Processing Systems}, 33.

\bibitem[{He et~al.(2020)He, Fan, Wu, Xie, and Girshick}]{he2020momentum}
He, K.; Fan, H.; Wu, Y.; Xie, S.; and Girshick, R. 2020.
\newblock Momentum contrast for unsupervised visual representation learning.
\newblock In \emph{Proceedings of the IEEE/CVF Conference on Computer Vision
  and Pattern Recognition}, 9729--9738.

\bibitem[{He et~al.(2016)He, Zhang, Ren, and Sun}]{he2016deep}
He, K.; Zhang, X.; Ren, S.; and Sun, J. 2016.
\newblock Deep residual learning for image recognition.
\newblock In \emph{Proceedings of the IEEE conference on computer vision and
  pattern recognition}, 770--778.

\bibitem[{Hsu, Qi, and Brown(2019)}]{hsu2019measuring}
Hsu, T.-M.~H.; Qi, H.; and Brown, M. 2019.
\newblock Measuring the effects of non-identical data distribution for
  federated visual classification.
\newblock arXiv:1909.06335.

\bibitem[{Huang et~al.(2021{\natexlab{a}})Huang, Li, Song, and
  Yang}]{huang2021fl}
Huang, B.; Li, X.; Song, Z.; and Yang, X. 2021{\natexlab{a}}.
\newblock FL-NTK: A Neural Tangent Kernel-based Framework for Federated
  Learning Analysis.
\newblock In \emph{International Conference on Machine Learning}, 4423--4434.
  PMLR.

\bibitem[{Huang et~al.(2021{\natexlab{b}})Huang, Chu, Zhou, Wang, Liu, Pei, and
  Zhang}]{huang2021personalized}
Huang, Y.; Chu, L.; Zhou, Z.; Wang, L.; Liu, J.; Pei, J.; and Zhang, Y.
  2021{\natexlab{b}}.
\newblock Personalized cross-silo federated learning on non-iid data.
\newblock In \emph{Proceedings of the AAAI Conference on Artificial
  Intelligence}, volume~35, 7865--7873.

\bibitem[{Jin et~al.(2020)Jin, Jiao, Qian, Zhang, Lu, and
  Wang}]{jin2020resource}
Jin, Y.; Jiao, L.; Qian, Z.; Zhang, S.; Lu, S.; and Wang, X. 2020.
\newblock Resource-efficient and convergence-preserving online participant
  selection in federated learning.
\newblock In \emph{2020 IEEE 40th International Conference on Distributed
  Computing Systems (ICDCS)}, 606--616. IEEE.

\bibitem[{Kairouz et~al.(2019)Kairouz, McMahan, Avent, Bellet, Bennis, Bhagoji,
  Bonawitz, Charles, Cormode, Cummings et~al.}]{kairouz2019advances}
Kairouz, P.; McMahan, H.~B.; Avent, B.; Bellet, A.; Bennis, M.; Bhagoji, A.~N.;
  Bonawitz, K.; Charles, Z.; Cormode, G.; Cummings, R.; et~al. 2019.
\newblock Advances and open problems in federated learning.
\newblock arXiv:1912.04977.

\bibitem[{Kaissis et~al.(2020)Kaissis, Makowski, R{\"u}ckert, and
  Braren}]{kaissis2020secure}
Kaissis, G.~A.; Makowski, M.~R.; R{\"u}ckert, D.; and Braren, R.~F. 2020.
\newblock Secure, privacy-preserving and federated machine learning in medical
  imaging.
\newblock \emph{Nature Machine Intelligence}, 2(6): 305--311.

\bibitem[{Karimireddy et~al.(2019)Karimireddy, Kale, Mohri, Reddi, Stich, and
  Suresh}]{karimireddy2019scaffold}
Karimireddy, S.~P.; Kale, S.; Mohri, M.; Reddi, S.~J.; Stich, S.~U.; and
  Suresh, A.~T. 2019.
\newblock SCAFFOLD: Stochastic Controlled Averaging for On-Device Federated
  Learning.

\bibitem[{Khosla et~al.(2020)Khosla, Teterwak, Wang, Sarna, Tian, Isola,
  Maschinot, Liu, and Krishnan}]{khosla2020supervised}
Khosla, P.; Teterwak, P.; Wang, C.; Sarna, A.; Tian, Y.; Isola, P.; Maschinot,
  A.; Liu, C.; and Krishnan, D. 2020.
\newblock Supervised Contrastive Learning.
\newblock \emph{Advances in Neural Information Processing Systems}, 33.

\bibitem[{Kumar et~al.(2021)Kumar, Khan, Kumar, Zakria, Golilarz, Zhang, Ting,
  Zheng, and Wang}]{kumar2021blockchain}
Kumar, R.; Khan, A.~A.; Kumar, J.; Zakria, A.; Golilarz, N.~A.; Zhang, S.;
  Ting, Y.; Zheng, C.; and Wang, W. 2021.
\newblock Blockchain-federated-learning and deep learning models for covid-19
  detection using ct imaging.
\newblock \emph{IEEE Sensors Journal}.

\bibitem[{Li et~al.(2020{\natexlab{a}})Li, Zhou, Xiong, and
  Hoi}]{li2020prototypical}
Li, J.; Zhou, P.; Xiong, C.; and Hoi, S. 2020{\natexlab{a}}.
\newblock Prototypical Contrastive Learning of Unsupervised Representations.
\newblock In \emph{International Conference on Learning Representations}.

\bibitem[{Li, He, and Song(2021)}]{li2021model}
Li, Q.; He, B.; and Song, D. 2021.
\newblock Model-Contrastive Federated Learning.
\newblock In \emph{Proceedings of the IEEE/CVF Conference on Computer Vision
  and Pattern Recognition}, 10713--10722.

\bibitem[{Li et~al.(2020{\natexlab{b}})Li, Sahu, Zaheer, Sanjabi, Talwalkar,
  and Smith}]{li2018federated}
Li, T.; Sahu, A.~K.; Zaheer, M.; Sanjabi, M.; Talwalkar, A.; and Smith, V.
  2020{\natexlab{b}}.
\newblock Federated Optimization in Heterogeneous Networks.
\newblock In \emph{Proceedings of Machine Learning and Systems}, volume~2,
  429--450.

\bibitem[{Liu et~al.(2020)Liu, Huang, Luo, Huang, Liu, Chen, Feng, Chen, Yu,
  and Yang}]{liu2020fedvision}
Liu, Y.; Huang, A.; Luo, Y.; Huang, H.; Liu, Y.; Chen, Y.; Feng, L.; Chen, T.;
  Yu, H.; and Yang, Q. 2020.
\newblock Fedvision: An online visual object detection platform powered by
  federated learning.
\newblock In \emph{Proceedings of the AAAI Conference on Artificial
  Intelligence}, volume~34, 13172--13179.

\bibitem[{Luping, Wei, and Bo(2019)}]{luping2019cmfl}
Luping, W.; Wei, W.; and Bo, L. 2019.
\newblock CMFL: Mitigating communication overhead for federated learning.
\newblock In \emph{2019 IEEE 39th International Conference on Distributed
  Computing Systems (ICDCS)}, 954--964. IEEE.

\bibitem[{McMahan et~al.(2017)McMahan, Moore, Ramage, Hampson, and
  y~Arcas}]{mcmahan2017communication}
McMahan, B.; Moore, E.; Ramage, D.; Hampson, S.; and y~Arcas, B.~A. 2017.
\newblock Communication-efficient learning of deep networks from decentralized
  data.
\newblock In \emph{Artificial intelligence and statistics}, 1273--1282. PMLR.

\bibitem[{Mohri, Sivek, and Suresh(2019)}]{mohri2019agnostic}
Mohri, M.; Sivek, G.; and Suresh, A.~T. 2019.
\newblock Agnostic federated learning.
\newblock In \emph{International Conference on Machine Learning}, 4615--4625.
  PMLR.

\bibitem[{Reisizadeh et~al.(2020)Reisizadeh, Farnia, Pedarsani, and
  Jadbabaie}]{reisizadeh2020robust}
Reisizadeh, A.; Farnia, F.; Pedarsani, R.; and Jadbabaie, A. 2020.
\newblock Robust Federated Learning: The Case of Affine Distribution Shifts.
\newblock In Larochelle, H.; Ranzato, M.; Hadsell, R.; Balcan, M.~F.; and Lin,
  H., eds., \emph{Advances in Neural Information Processing Systems},
  volume~33, 21554--21565. Curran Associates, Inc.

\bibitem[{Sattler et~al.(2019)Sattler, Wiedemann, M{\"u}ller, and
  Samek}]{sattler2019robust}
Sattler, F.; Wiedemann, S.; M{\"u}ller, K.-R.; and Samek, W. 2019.
\newblock Robust and communication-efficient federated learning from non-iid
  data.
\newblock \emph{IEEE transactions on neural networks and learning systems},
  31(9): 3400--3413.

\bibitem[{Snell, Swersky, and Zemel(2017)}]{snell2017prototypical}
Snell, J.; Swersky, K.; and Zemel, R. 2017.
\newblock Prototypical networks for few-shot learning.
\newblock In \emph{Proceedings of the 31st International Conference on Neural
  Information Processing Systems}, 4080--4090.

\bibitem[{T~Dinh, Tran, and Nguyen(2020)}]{t2020personalized}
T~Dinh, C.; Tran, N.; and Nguyen, T.~D. 2020.
\newblock Personalized Federated Learning with Moreau Envelopes.
\newblock \emph{Advances in Neural Information Processing Systems}, 33.

\bibitem[{Truex et~al.(2019)Truex, Baracaldo, Anwar, Steinke, Ludwig, Zhang,
  and Zhou}]{truex2019hybrid}
Truex, S.; Baracaldo, N.; Anwar, A.; Steinke, T.; Ludwig, H.; Zhang, R.; and
  Zhou, Y. 2019.
\newblock A hybrid approach to privacy-preserving federated learning.
\newblock In \emph{Proceedings of the 12th ACM Workshop on Artificial
  Intelligence and Security}, 1--11.

\bibitem[{van Berlo, Saeed, and Ozcelebi(2020)}]{van2020towards}
van Berlo, B.; Saeed, A.; and Ozcelebi, T. 2020.
\newblock Towards federated unsupervised representation learning.
\newblock In \emph{Proceedings of the Third ACM International Workshop on Edge
  Systems, Analytics and Networking}, 31--36.

\bibitem[{van~der Maaten and Hinton(2008)}]{van2008visualizing}
van~der Maaten, L.; and Hinton, G. 2008.
\newblock Visualizing Data using t-SNE.
\newblock \emph{Journal of Machine Learning Research}, 9: 2579--2605.

\bibitem[{Wang et~al.(2020{\natexlab{a}})Wang, Yurochkin, Sun, Papailiopoulos,
  and Khazaeni}]{Wang2020Federated}
Wang, H.; Yurochkin, M.; Sun, Y.; Papailiopoulos, D.; and Khazaeni, Y.
  2020{\natexlab{a}}.
\newblock Federated Learning with Matched Averaging.
\newblock In \emph{International Conference on Learning Representations}.

\bibitem[{Wang et~al.(2020{\natexlab{b}})Wang, Liu, Liang, Joshi, and
  Poor}]{NEURIPS2020_564127c0}
Wang, J.; Liu, Q.; Liang, H.; Joshi, G.; and Poor, H.~V. 2020{\natexlab{b}}.
\newblock Tackling the Objective Inconsistency Problem in Heterogeneous
  Federated Optimization.
\newblock In Larochelle, H.; Ranzato, M.; Hadsell, R.; Balcan, M.~F.; and Lin,
  H., eds., \emph{Advances in Neural Information Processing Systems},
  volume~33, 7611--7623. Curran Associates, Inc.

\bibitem[{Wang et~al.(2021)Wang, Han, Wei, Zhang, and
  Wang}]{wang2021contrastive}
Wang, P.; Han, K.; Wei, X.-S.; Zhang, L.; and Wang, L. 2021.
\newblock Contrastive Learning based Hybrid Networks for Long-Tailed Image
  Classification.
\newblock In \emph{Proceedings of the IEEE/CVF Conference on Computer Vision
  and Pattern Recognition}, 943--952.

\bibitem[{Wang et~al.(2019)Wang, Song, Zhang, Song, Wang, and
  Qi}]{wang2019beyond}
Wang, Z.; Song, M.; Zhang, Z.; Song, Y.; Wang, Q.; and Qi, H. 2019.
\newblock Beyond inferring class representatives: User-level privacy leakage
  from federated learning.
\newblock In \emph{IEEE INFOCOM 2019-IEEE Conference on Computer
  Communications}, 2512--2520. IEEE.

\bibitem[{Yurochkin et~al.(2019)Yurochkin, Agarwal, Ghosh, Greenewald, Hoang,
  and Khazaeni}]{yurochkin2019bayesian}
Yurochkin, M.; Agarwal, M.; Ghosh, S.; Greenewald, K.; Hoang, N.; and Khazaeni,
  Y. 2019.
\newblock Bayesian nonparametric federated learning of neural networks.
\newblock In \emph{International Conference on Machine Learning}, 7252--7261.
  PMLR.

\bibitem[{Zhang et~al.(2020)Zhang, Kuang, You, Shen, Xiao, Zhang, Wu, Zhuang,
  and Li}]{zhang2020federated}
Zhang, F.; Kuang, K.; You, Z.; Shen, T.; Xiao, J.; Zhang, Y.; Wu, C.; Zhuang,
  Y.; and Li, X. 2020.
\newblock Federated unsupervised representation learning.
\newblock arXiv:2010.08982.

\bibitem[{{Zhao} et~al.(2018){Zhao}, {Li}, {Lai}, {Suda}, {Civin}, and
  {Chandra}}]{zhao2018federated}
{Zhao}, Y.; {Li}, M.; {Lai}, L.; {Suda}, N.; {Civin}, D.; and {Chandra}, V.
  2018.
\newblock Federated Learning with Non-IID Data.
\newblock arXiv:1806.00582.

\bibitem[{Zhou et~al.(2020)Zhou, Cui, Wei, and Chen}]{zhou2020bbn}
Zhou, B.; Cui, Q.; Wei, X.-S.; and Chen, Z.-M. 2020.
\newblock Bbn: Bilateral-branch network with cumulative learning for
  long-tailed visual recognition.
\newblock In \emph{Proceedings of the IEEE/CVF Conference on Computer Vision
  and Pattern Recognition}, 9719--9728.

\end{thebibliography}

\end{document}